\pdfoutput=1

\documentclass[11pt]{article}

\usepackage{acl}

\usepackage{times}
\usepackage{latexsym}

\usepackage[T1]{fontenc}

\usepackage[utf8]{inputenc}

\usepackage{microtype}

\usepackage{inconsolata}

\usepackage{amsfonts}
\usepackage{amsmath}
\usepackage{booktabs}
\usepackage{graphicx}
\usepackage{multirow}
\usepackage{pifont}

\newcommand{\sd}[1]{\textsubscript{\,\textcolor{gray}{#1}}}

\title{Vocabulary-level Memory Efficiency for Language Model Fine-tuning}

\author{Miles Williams \and Nikolaos Aletras \\
  University of Sheffield \\
  \texttt{\{mwilliams15, n.aletras\}@sheffield.ac.uk}}

\begin{document}
\maketitle
\begin{abstract}
The extensive memory footprint of language model (LM) fine-tuning poses a challenge for both researchers and practitioners. LMs use an embedding matrix to represent extensive vocabularies, forming a substantial proportion of the model parameters. While previous work towards memory-efficient fine-tuning has focused on minimizing the number of trainable parameters, reducing the memory footprint of the embedding matrix has yet to be explored. We first demonstrate that a significant proportion of the vocabulary remains unused during fine-tuning. We then propose a simple yet effective approach that leverages this finding to minimize memory usage. We show that our approach provides substantial reductions in memory usage across a wide range of models and tasks. Notably, our approach does not impact downstream task performance, while allowing more efficient use of computational resources.\footnote{\url{https://github.com/mlsw/partial-embedding-matrix-adaptation}}
\end{abstract}

\section{Introduction}

Language models (LMs) \citep{chung-etal-2022-scaling, touvron-etal-2023-llama, warner-etal-2024-smarter} form the foundation of contemporary natural language processing (NLP), however they require extensive computational resources to train \citep{kaplan-etal-2020-scaling, hoffmann-etal-2022-empirical}. This is contrary to the democratization of NLP, exacerbating economic inequalities and hindering inclusivity \citep{schwartz-etal-2020-green, weidinger-etal-2022-taxonomy}. Consequently, there is a growing focus towards developing efficient methods for LM training and fine-tuning \citep{treviso-etal-2023-efficient, lialin-etal-2023-scaling}.

The memory footprint of LMs is a major challenge for their application. Storing model parameters requires extensive amounts of memory, constraining the size and architecture of the model \citep{paleyes-etal-2022-challenges}. This problem is especially prominent during training as gradients and optimizer states must also be retained \citep{kingma-ba-2015-adam}. This can be problematic when using consumer hardware or facing an academic budget \citep{izsak-etal-2021-train, ciosici-derczynski-2022-training}.

LMs ordinarily use fixed vocabularies to derive vector representations from text, known as word embeddings. Each element of the vocabulary has a corresponding word embedding, which collectively form an embedding matrix within the LM. The size of the embedding matrix scales with both the vocabulary size and embedding dimension, comprising a substantial proportion of the model parameters (Table~\ref{tab:vocabulary_sizes}, Appendix~\ref{app:vocabulary_sizes}). This proportion is usually even greater for multilingual LMs, which benefit from larger vocabularies \citep{conneau-etal-2020-unsupervised, liang-etal-2023-xlm}. However, we hypothesize that a significant proportion of LM vocabulary remains unused during fine-tuning on many downstream tasks.

\begin{figure}[t]
    \centering
    \includegraphics[scale=0.7]{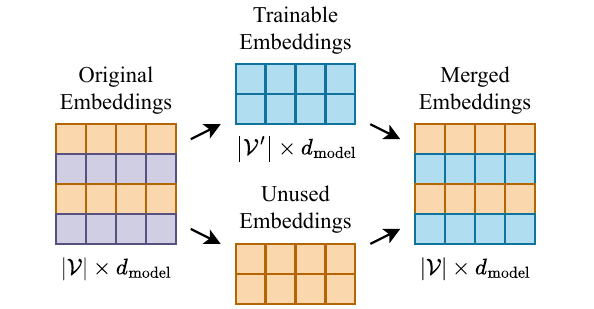}
    \caption{Memory-efficient language model fine-tuning with Partial Embedding Matrix Adaptation (PEMA).}
    \label{fig:diagram}
\end{figure}

In this paper, we first demonstrate that our hypothesis holds for a variety of downstream tasks, with only a small subset of vocabulary used. We then propose a method to reduce memory usage during fine-tuning by excluding unused embeddings. Finally, we empirically demonstrate the memory savings from our approach across a range of models and tasks. Notably, our approach does not impact downstream task performance and is orthogonal to many existing LM memory efficiency techniques.

\section{Related Work}
\label{sec:related_work}

\paragraph{Tokenization.}

Transformer LMs \citep{vaswani-etal-2017-attention} typically adopt subword tokenization \citep{schuster-nakajima-2012-japanese, sennrich-etal-2016-neural} to encode text using a finite vocabulary. The use of large subword vocabularies enables improved task performance \citep{galle-2019-investigating}, inference efficiency \citep{tay-etal-2022-efficient}, and multilingual performance \citep{liang-etal-2023-xlm}. Conversely, character or byte level tokenization can be used \citep{clark-etal-2022-canine, xue-etal-2022-byt5}, reducing the size of the embedding matrix at the cost of increasing the sequence length.

\paragraph{Reducing embedding parameters.}

To reduce the size of the embedding matrix, LMs can be trained with embedding factorization \citep{sun-etal-2020-mobilebert, lan-etal-2020-albert}, albeit with slightly lower task performance. Alternatively, embeddings can be generated from hash functions \citep{sankar-etal-2021-proformer, xue-aletras-2022-hashformers, cohn-etal-2023-eelbert}, although this may harm performance due to the many-to-one mapping from tokens to embeddings.

\paragraph{Multilingual vocabulary trimming.}

The closest work to our own is \citet{abdaoui-etal-2020-load}, which creates smaller multilingual LMs by permanently reducing the number of supported languages. This can harm performance as the removed vocabulary may later be required for a downstream task. Moreover, selecting which vocabulary to remove requires the computationally expensive processing of a large corpus. \citet{ushio-etal-2023-efficient} further examine the performance impact of permanently removing LM vocabulary either before or after fine-tuning. However, the same fundamental limitations persist.

\paragraph{Parameter-efficient fine-tuning.}

PEFT methods, such as adapters \citep{houlsby-etal-2019-parameter}, soft prompts \citep{lester-etal-2021-power, li-liang-2021-prefix}, ladder side-tuning \citep{sung-etal-2022-lst}, and low-rank adaptation \citep{hu-etal-2022-lora}, effectively adapt LMs by fine-tuning only a small number of parameters. However, these methods still require all LM parameters to be held in accelerator memory.

\paragraph{Offloading.}

To minimize accelerator (e.g. GPU) memory usage, LM parameters can be held in separate (e.g. CPU) memory until needed \citep{pudipeddi-etal-2020-training, ren-etal-2021-zero}. However, this approach substantially increases inference latency.

\paragraph{Model compression.} In Appendix~\ref{app:language_model_compression}, we discuss a variety of orthogonal LM compression methods, such as quantization, pruning, and distillation.

\section{Vocabulary Usage Analysis}
\label{sec:vocabulary_usage_analysis}

\begin{figure}[t]
\centering
\includegraphics{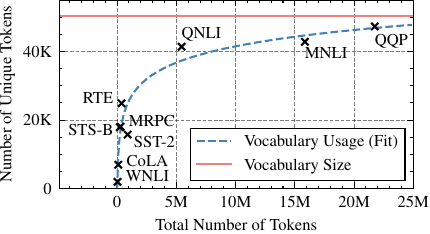}
\caption{The trend in vocabulary use for the datasets in GLUE when using the vocabulary from GPT-2.}
\label{fig:glue_datatasets}
\end{figure}

\begin{table}[t]
\scriptsize
\centering
\begin{tabular}{cl}
\toprule
\# & Token \\ \midrule
49,990 & \texttt{natureconservancy} \\
50,072 & \texttt{;;;;;;;;;;;;} \\
50,160 & \texttt{PsyNetMessage} \\
50,174 & \texttt{rawdownloadcloneembedreportprint} \\
50,243 & \texttt{SolidGoldMagikarp} \\ \bottomrule
\end{tabular}
\caption{Five examples of tokens from the GPT-2 vocabulary that do not occur within English Wikipedia.}
\label{tab:anomalous_tokens}
\end{table}

To empirically assess the level of vocabulary usage during fine-tuning, we first examine the popular GLUE benchmark \citep{wang-etal-2019-glue}. This comprises a series of tasks that are varied in both size and domain (Appendix~\ref{app:datasets}). For tokenization, we use the subword vocabulary from GPT-2, which was later adopted by models including RoBERTa \citep{liu-etal-2019-roberta}, BART \citep{lewis-etal-2020-bart}, GPT-3 \citep{brown-etal-2020-language}, and OPT \citep{zhang-etal-2022-opt}.

Figure~\ref{fig:glue_datatasets} illustrates the relationship between unique tokens and total tokens in each of the GLUE datasets. Notably, six out of nine datasets fail to use more than half of the vocabulary. Moreover, the smallest dataset, WNLI, uses less than 4\%. Interestingly, we observe that the GLUE datasets follow a trend resembling Heaps' Law \citep{heaps-1978-information}. This states that as the size of a corpus grows, there are diminishing gains in new vocabulary. However, our use of a finite subword vocabulary means that the trend is asymptotic to the vocabulary size.

Separately, the statistical construction of subword vocabularies can reflect anomalies in their training data, creating tokens that may never be used. To examine the extent of the issue, we identify such tokens by evaluating a processed dump of English Wikipedia, comprising over 20GB of text. Peculiarly, we identify nearly 200 anomalous tokens without a single occurrence (see Table~\ref{tab:anomalous_tokens}).\footnote{We refer readers interested in such anomalous tokens to \citet{rumbelow-watkins-2023-solidgoldmagikarp} and \citet{land-bartolo-2024-fishing}.}

\section{Partial Embedding Matrix Adaptation}

Our empirical analysis (Section~\ref{sec:vocabulary_usage_analysis}) suggests that many fine-tuning datasets only use a fraction of LM vocabulary. We leverage this insight to propose Partial Embedding Matrix Adaptation (PEMA), a method that achieves substantial memory savings by selecting only the minimum subset of word embeddings needed for fine-tuning. Notably, this does not impact task performance, as unused word embeddings are not updated during backpropagation.

\paragraph{Preliminaries.}

Let each token in the vocabulary $\{w_1, \dots, w_k\}$ be denoted by a unique integer $i$ such that $\mathcal{V} = \{i \in \mathbb{N} \mid i \le k \}$. The embedding matrix $E \in \mathbb{R}^{|\mathcal{V}|\times d}$ is then used to project each token to a corresponding $d$-dimensional vector.

\paragraph{Before fine-tuning.}

Suppose we have fine-tuning dataset $D \in \mathcal{V}^{m \times n}$ where $m$ is the number of examples and $n$ is the length of each example. We compute the partial vocabulary $\mathcal{V}' \subset \mathcal{V}$ consisting of \emph{only} the tokens in $D$. As the elements of $\mathcal{V}'$ are not necessarily consecutive integers, we define an arbitrary mapping $f \colon \mathcal{V}' \rightarrow \{i \in \mathbb{N} \mid i \le |\mathcal{V}'|\}$. We then construct the partial embedding matrix $E' \in \mathbb{R}^{|\mathcal{V}'|\times d}$ with entries $E'[:, f(i)] = E[:, i]$ for all $i \in \mathcal{V}'$. That is, $E'$ retains only embedding vectors corresponding to tokens in $\mathcal{V}'$. To adapt $D$ for the partial vocabulary $\mathcal{V}'$, we create an intermediary dataset $D'$ where each entry $D'[i,j] = f(D[i,j])$. Finally, we use $D'$ and $E'$ in place of $D$ and $E$.

\paragraph{After fine-tuning.}

Following fine-tuning, our partial embedding matrix $E'$ holds the newly learned embeddings for the partial vocabulary. However, we do not wish to keep only the partial vocabulary, as this would limit future use of the model (i.e. tasks with different vocabulary). Therefore, we merge the newly learned embeddings into the original embedding matrix (stored on-disk). More formally, we update $E$ such that $E[:, f^{-1}(i)] = E'[:, i]$ for all $i \in \mathcal{V}'$. This ensures that the model remains structurally identical, with embeddings for the complete vocabulary.

\section{Experimental Setup}

\paragraph{Datasets.}

To offer a fair selection of datasets, we follow existing PEFT literature \citep{houlsby-etal-2019-parameter, hu-etal-2022-lora, sung-etal-2022-lst, zhang-etal-2023-adaptive-budget} and focus our evaluation on the popular GLUE benchmark. We additionally employ XNLI \citep{conneau-etal-2018-xnli} to assess the performance of our approach with multilingual data. Complete data sources and implementation details are listed in Appendix~\ref{app:datasets} and Appendix~\ref{app:implementation}, respectively.

\paragraph{Models.}

Similarly, we select a variety of popular models used in existing work. However, we place an emphasis on having a variety of vocabularies (Table~\ref{tab:vocabulary_sizes}, Appendix~\ref{app:vocabulary_sizes}). For monolingual models, we use BERT \citep{devlin-etal-2019-bert}, RoBERTa \citep{liu-etal-2019-roberta}, and DeBERTaV3 \citep{he-etal-2023-deberta}. For multilingual models, we use mBERT \citep{devlin-etal-2019-bert}, XLM-RoBERTa \citep{conneau-etal-2020-unsupervised}, and XLM-V \citep{liang-etal-2023-xlm}. To evaluate the performance of distilled models, we also use the available distilled counterparts: DistilBERT, DistilRoBERTa, and DistilmBERT \citep{sanh-etal-2019-distilbert}. For a fair comparison between models, we consistently select the base size ($d_\text{model}=768$).

\paragraph{Memory efficiency metrics.}

Following convention in the PEFT literature \citep{houlsby-etal-2019-parameter, hu-etal-2022-lora, ben-zaken-etal-2022-bitfit}, we report memory efficiency in terms of model parameters. This is advantageous as it avoids confounding factors such as weight precision, optimizer choice, software implementation, and batch size.

\section{Results}

\begin{table*}[t]
\scriptsize
\centering
\begin{tabular}{lrrrrrrrrrr}
\toprule
Model & \multicolumn{1}{c}{CoLA} & \multicolumn{1}{c}{MNLI} & \multicolumn{1}{c}{MRPC} & \multicolumn{1}{c}{QNLI} & \multicolumn{1}{c}{QQP} & \multicolumn{1}{c}{RTE} & \multicolumn{1}{c}{SST-2} & \multicolumn{1}{c}{STS-B} & \multicolumn{1}{c}{WNLI} & \multicolumn{1}{c}{Mean} \\
\midrule
\multicolumn{11}{c}{Reduction in Embedding Parameters (\%)} \\
\midrule
DistilBERT & 80.1 & 14.8 & 54.9 & 13.1 & 11.5 & 41.5 & 57.9 & 57.2 & 94.3 & 47.3 \\
DistilRoBERTa & 86.1 & 14.8 & 64.0 & 17.7 & 5.9 & 51.6 & 68.6 & 64.4 & 96.0 & 52.1 \\
DistilmBERT & 94.9 & 76.9 & 88.2 & 73.8 & 72.7 & 85.0 & 91.9 & 88.8 & 98.4 & 85.6 \\
\cmidrule{1-11}
BERT & 80.1 & 14.8 & 54.9 & 13.1 & 11.5 & 41.5 & 57.9 & 57.2 & 94.3 & 47.3 \\
RoBERTa & 86.1 & 14.8 & 64.0 & 17.7 & 5.9 & 51.6 & 68.6 & 64.4 & 96.0 & 52.1 \\
DeBERTaV3 & 95.0 & 44.3 & 85.7 & 47.1 & 28.5 & 79.0 & 87.5 & 85.9 & 98.6 & 72.4 \\
\cmidrule{1-11}
mBERT & 94.9 & 76.9 & 88.2 & 73.8 & 72.7 & 85.0 & 91.9 & 88.8 & 98.4 & 85.6 \\
XLM-RoBERTa & 97.8 & 88.8 & 94.9 & 87.6 & 85.4 & 93.3 & 96.3 & 94.9 & 99.3 & 93.1 \\
XLM-V & 99.3 & 93.2 & 98.0 & 92.8 & 90.5 & 97.1 & 98.3 & 98.0 & 99.8 & 96.3 \\
\midrule
\multicolumn{11}{c}{Reduction in Model Parameters (\%)} \\
\midrule
DistilBERT & 28.0 & 5.2 & 19.2 & 4.6 & 4.0 & 14.5 & 20.3 & 20.0 & 33.0 & 16.5 \\
DistilRoBERTa & 40.5 & 7.0 & 30.1 & 8.3 & 2.8 & 24.3 & 32.3 & 30.3 & 45.1 & 24.5 \\
DistilmBERT & 64.4 & 52.2 & 59.9 & 50.1 & 49.3 & 57.7 & 62.3 & 60.2 & 66.8 & 58.1 \\
\cmidrule{1-11}
BERT & 17.1 & 3.2 & 11.8 & 2.8 & 2.5 & 8.9 & 12.4 & 12.2 & 20.2 & 10.1 \\
RoBERTa & 26.7 & 4.6 & 19.8 & 5.5 & 1.8 & 16.0 & 21.2 & 19.9 & 29.7 & 16.1 \\
DeBERTaV3 & 50.7 & 23.6 & 45.7 & 25.1 & 15.2 & 42.1 & 46.7 & 45.8 & 52.6 & 38.6 \\
\cmidrule{1-11}
mBERT & 49.0 & 39.7 & 45.5 & 38.1 & 37.5 & 43.9 & 47.4 & 45.8 & 50.8 & 44.2 \\
XLM-RoBERTa & 67.5 & 61.3 & 65.5 & 60.5 & 59.0 & 64.4 & 66.5 & 65.5 & 68.5 & 64.3 \\
XLM-V & 88.3 & 82.9 & 87.2 & 82.6 & 80.5 & 86.4 & 87.5 & 87.2 & 88.8 & 85.7 \\
\bottomrule
\end{tabular}
\caption{The reduction in embedding and model parameters (\%) for each model across the GLUE benchmark.}
\label{tab:glue_models}
\end{table*}

\begin{table*}[t]
\scriptsize
\centering
\begin{tabular}{lrrrrrrrrrr}
\toprule
Size & \multicolumn{1}{c}{CoLA} & \multicolumn{1}{c}{MNLI} & \multicolumn{1}{c}{MRPC} & \multicolumn{1}{c}{QNLI} & \multicolumn{1}{c}{QQP} & \multicolumn{1}{c}{RTE} & \multicolumn{1}{c}{SST-2} & \multicolumn{1}{c}{STS-B} & \multicolumn{1}{c}{WNLI} & \multicolumn{1}{c}{Mean} \\
\midrule
XSmall & 46.7 & 21.8 & 42.2 & 23.2 & 14.0 & 38.8 & 43.1 & 42.3 & 48.5 & 35.6 \\
Small & 93.4 & 43.6 & 84.3 & 46.3 & 28.0 & 77.7 & 86.1 & 84.5 & 97.0 & 71.2 \\
Base & 93.4 & 43.6 & 84.3 & 46.3 & 28.0 & 77.7 & 86.1 & 84.5 & 97.0 & 71.2 \\
Large & 124.6 & 58.1 & 112.4 & 61.8 & 37.3 & 103.6 & 114.8 & 112.7 & 129.4 & 95.0 \\
\bottomrule
\end{tabular}
\caption{The reduction in model parameters (millions) for each size of DeBERTaV3 across the GLUE benchmark.}
\label{tab:glue_deberta_models}
\end{table*}

\paragraph{Larger vocabularies see more memory savings.}

Table~\ref{tab:glue_models} presents the reduction in parameters for each model across the GLUE benchmark. Following our expectations from Section~\ref{sec:vocabulary_usage_analysis}, we generally observe that as vocabulary sizes increase (Table~\ref{tab:vocabulary_sizes}, Appendix~\ref{app:vocabulary_sizes}), so do the potential memory savings. For example, an average reduction in embedding parameters of 47.3\% is achieved for BERT, 52.1\% for RoBERTa, and 72.4\% for DeBERTaV3. 

\paragraph{Memory savings vary between datasets.}

In line with our expectations from Section \ref{sec:vocabulary_usage_analysis}, the memory savings vary substantially between datasets. For BERT, the embedding matrix can be reduced by 94.3\% for the smallest dataset (WNLI), yet only 11.5\% for the largest (QQP). We demonstrate that downstream task performance remains consistent across models and datasets in Appendix \ref{app:glue_results}.

\paragraph{Distilled models substantially benefit.}

Considering the distilled models, we observe that they all achieve an identical reduction in embedding parameters to their original counterparts. This is because they use the same vocabulary and embedding size \citep{sanh-etal-2019-distilbert}. However, they offer substantially higher overall savings, as there are fewer parameters allocated to the transformer layers.

\paragraph{Memory savings scale with model size.}

Table~\ref{tab:glue_deberta_models} presents the reduction in model parameters for each model from the DeBERTaV3 family. We observe that this reduction continues to increase with model size. On average, the extra small size is reduced by 35.6M parameters, while the large size is reduced by 95.0M parameters. Although the same fixed-size vocabulary is shared across models, the embedding dimension continues to grow (Table~\ref{tab:deberta_sizes}, Appendix~\ref{app:vocabulary_sizes}), offering further memory savings. The exception to this is the small and base sizes, where the only difference is the number of layers.

\paragraph{Multilingual models achieve extreme savings.}

Unsurprisingly, multilingual models demonstrate extreme memory savings across the monolingual GLUE benchmark. On average, a reduction in model parameters of 44.2\% is achieved for mBERT, 64.3\% for XLM-RoBERTa, and 85.7\% for XLM-V. Table~\ref{tab:xnli_models} presents the reduction in parameters for the multilingual models when fine-tuning on different subsets of XNLI. Even when fine-tuning on all fifteen languages, these models still demonstrate substantial memory savings from 23.0\% to 58.4\%.

\begin{table}[t]
\scriptsize
\centering
\begin{tabular}{lrrrr}
\toprule
Model & \multicolumn{1}{c}{en} & \multicolumn{1}{c}{en-de} & \multicolumn{1}{c}{en-zh} & \multicolumn{1}{c}{All} \\
\midrule
\multicolumn{5}{c}{Reduction in Embedding Parameters (\%)} \\
\midrule
DistilmBERT & 77.1 & 71.7 & 73.0 & 44.6 \\
mBERT & 77.1 & 71.7 & 73.0 & 44.6 \\
XLM-RoBERTa & 89.2 & 86.0 & 84.4 & 56.9 \\
XLM-V & 93.6 & 90.0 & 90.0 & 65.7 \\
\midrule
\multicolumn{5}{c}{Reduction in Model Parameters (\%)} \\
\midrule
DistilmBERT & 52.3 & 48.6 & 49.6 & 30.3 \\
mBERT & 39.8 & 37.0 & 37.7 & 23.0 \\
XLM-RoBERTa & 61.6 & 59.4 & 58.3 & 39.3 \\
XLM-V & 83.2 & 80.0 & 80.0 & 58.4 \\
\bottomrule
\end{tabular}
\caption{The reduction in parameters across different subsets of XNLI, in addition to all fifteen languages.}
\label{tab:xnli_models}
\end{table}

\section{Conclusion}

In this paper, we identified that many fine-tuning datasets do not use the majority of LM vocabulary. We then proposed Partial Embedding Matrix Adaptation (PEMA), a simple yet effective approach to minimize LM memory use during fine-tuning, that is orthogonal to many existing methods. Finally, we empirically demonstrated that our approach offers substantial memory savings across a variety of popular tasks and models, without compromising performance. As future work, we are interested in adapting our approach for the output embedding matrix to offer further memory savings.

\section*{Limitations}

Processing the fine-tuning dataset to assess vocabulary usage incurs a runtime cost. However, we observe that this cost is negligible. We provide a detailed analysis of this matter in Appendix~\ref{app:runtime_impact}.

\section*{Ethical Considerations}

Our approach improves the memory efficiency of LM fine-tuning, therefore facilitating the use of less powerful hardware. Although we hope that this can reduce the environmental footprint of LM fine-tuning, we acknowledge that it could be used to support the fine-tuning of even larger LMs. We also recognize the dual-use nature of LMs and concede that efforts towards improving efficiency, including our own, can lower the barrier to entry for their misuse \citep{weidinger-etal-2022-taxonomy}.

\section*{Acknowledgments}

We are sincerely grateful to Nafise Sadat Moosavi, Huiyin Xue, Atsuki Yamaguchi, and the anonymous reviewers for their invaluable feedback. MW is supported by the Centre for Doctoral Training in Speech and Language Technologies (SLT) and their Applications funded by UK Research and Innovation grant EP/S023062/1. 

\bibliography{anthology,custom}

\clearpage

\appendix

\section{Language Model Vocabulary Sizes}
\label{app:vocabulary_sizes}

Table~\ref{tab:vocabulary_sizes} presents the vocabulary sizes ($|\mathcal{V}|$) for the models used in our experiments, as identified by the Hugging Face Hub. We also report the number of embedding parameters ($N_\text{emb}$), the number of model parameters ($N$), and the overall proportion of embedding parameters ($N_\text{emb}/N$). These metrics are also presented in Table~\ref{tab:deberta_sizes} for each size of DeBERTa, in addition to model hyperparameters.

\section{Language Model Compression}
\label{app:language_model_compression}

Supplementary to our discussion of related work (Section~\ref{sec:related_work}), we additionally discuss the relation to variety of popular LM compression approaches. We emphasize that these methods are orthogonal to our proposed approach.

\paragraph{Knowledge distillation.}

Knowledge distillation \citep{hinton-etal-2015-distilling} aims to achieve comparable performance by training a smaller model using the predictions from a larger model. This approach has been successfully applied to LMs \citep{sanh-etal-2019-distilbert, sun-etal-2020-mobilebert}. It can also be used to train models with a smaller vocabulary than the original \citep{zhao-etal-2021-extremely, singh-lefever-2022-student}.

\paragraph{Pruning.}

Neural network pruning \citep{lecun-etal-1989-optimal} seeks to remove redundant weights while preserving performance. Existing approaches focus on pruning the linear and attention weights in LMs \citep{sanh-etal-2020-movement, kurtic-etal-2022-optimal, frantar-alistarh-2023-sparsegpt}. However, pruning the embedding matrix is widely avoided, as it can substantially harm performance \citep{kurtic-etal-2024-how}.

\paragraph{Quantization.}

The aim of quantization is to represent neural network weights using lower precision, therefore reducing computational costs. Recent LM quantization efforts generally focus on quantizing the linear layers \citep{dettmers-etal-2022-gpt, yao-etal-2022-zeroquant, frantar-etal-2023-optq}. The embedding matrix can also be quantized \citep{zafrir-etal-2019-q8bert, bondarenko-etal-2021-understanding}, although \citet{shen-etal-2020-qbert} find that it is more sensitive to quantization.

\begin{table}[t]
\centering
\scriptsize
\begin{tabular}{lrrrr}
\toprule
Model & \multicolumn{1}{c}{$|\mathcal{V}|$} & \multicolumn{1}{c}{$N_{\text{emb}}$} & \multicolumn{1}{c}{$N$} & \multicolumn{1}{c}{$N_{\text{emb}}/N$} \\
\midrule
DistilBERT & 28,996 & 22.3M & 65.8M & 33.9\% \\
DistilRoBERTa & 50,265 & 38.6M & 82.1M & 47.0\% \\
DistilmBERT & 119,547 & 91.8M & 135.3M & 67.8\% \\
\cmidrule{1-5}
BERT & 28,996 & 22.3M & 108.3M & 20.6\% \\
RoBERTa & 50,265 & 38.6M & 124.6M & 31.0\% \\
DeBERTaV3 & 128,100 & 98.4M & 184.4M & 53.3\% \\
\cmidrule{1-5}
mBERT & 119,547 & 91.8M & 177.9M & 51.6\% \\
XLM-RoBERTa & 250,002 & 192.0M & 278.0M & 69.1\% \\
XLM-V & 901,629 & 692.5M & 778.5M & 88.9\% \\
\bottomrule
\end{tabular}
\caption{The vocabulary size and allocation of parameters for each of the models used in our experiments. In all cases, we select the base model size ($d_\text{model}=768$).}
\label{tab:vocabulary_sizes}
\end{table}

\begin{table}[t]
\centering
\scriptsize
\begin{tabular}{lrrrrrr}
\toprule
Size & \multicolumn{1}{c}{$l$} & \multicolumn{1}{c}{$h$} & \multicolumn{1}{c}{$d_{\text{model}}$} & \multicolumn{1}{c}{$N_{\text{emb}}$} & \multicolumn{1}{c}{$N$} & \multicolumn{1}{c}{$N_{\text{emb}}/N$} \\
\midrule
XSmall & 12 & 6 & 384 & 49.2M & 70.8M & 69.4\% \\
Small & 6 & 12 & 768 & 98.4M & 141.9M & 69.3\% \\
Base & 12 & 12 & 768 & 98.4M & 184.4M & 53.3\% \\
Large & 24 & 16 & 1024 & 131.2M & 435.1M & 30.2\% \\
\bottomrule
\end{tabular}
\caption{The DeBERTaV3 \citep{he-etal-2023-deberta} family of models. Columns $l$, $h$, and $d_{\textrm{model}}$ show the number of hidden layers, number of attention heads, and hidden embedding size, respectively.
}
\label{tab:deberta_sizes}
\end{table}

\section{Datasets}
\label{app:datasets}

In all cases, we use the publicly available version of each dataset available from Hugging Face \citep{lhoest-etal-2021-datasets}. The GLUE benchmark comprises a diverse range of tasks, including linguistic acceptability (CoLA, \citealt{warstadt-etal-2019-neural}), sentiment analysis (SST-2, \citealt{socher-etal-2013-recursive}), paraphrasing/sentence similarity (MRPC, \citealt{dolan-brockett-2005-automatically}; STS-B, \citealt{cer-etal-2017-semeval}; QQP, \citealt{iyer-etal-2017-first}), and natural language inference (RTE, \citealt{dagan-etal-2006-pascal}; WNLI, \citealt{levesque-etal-2012-winograd}; QNLI, \citealt{rajpurkar-etal-2016-squad}; MNLI, \citealt{williams-etal-2018-broad}). The number of examples per split in each dataset are listed in Table \ref{tab:glue_splits}. The XNLI dataset \citep{conneau-etal-2018-xnli} extends MNLI to 15 languages: Arabic, Bulgarian, Chinese, English, French, German, Greek, Hindi, Russian, Spanish, Swahili, Thai, Turkish, Vietnamese, and Urdu.

\section{Implementation \& Hardware}
\label{app:implementation}

We implement our experiments using PyTorch \citep{paszke-etal-2019-pytorch}, Hugging Face Transformers \citep{wolf-etal-2020-transformers} and Hugging Face Datasets \citep{lhoest-etal-2021-datasets}. Since downstream task performance is not relevant to this study, we do not perform hyperparameter tuning. Instead, we broadly follow the hyperparameters from \citet{devlin-etal-2019-bert}, listed in Table~\ref{tab:hyperparameters}.

We fine-tune all models using a single NVIDIA Tesla V100 (SXM2 32GB) GPU and Intel Xeon Gold 6138 CPU. For consistency, each model type is evaluated on the same physical hardware.

\section{Fine-tuning on GLUE}
\label{app:glue_results}

Table~\ref{tab:glue_performance} presents the task performance for each model across the GLUE benchmark. We observe that the performance is largely identical, although there are occasional fluctuations where PEMA performs fractionally better or worse than the baseline. Finally, we note that XLM-RoBERTa and XLM-V both demonstrate very low performance on CoLA, although this issue has also been observed in other studies, e.g. \citet{zhou-etal-2023-predictive}.

\section{Runtime Impact}
\label{app:runtime_impact}

Table~\ref{tab:runtime_impact} presents the mean duration and standard deviation of applying PEMA to RoBERTa and the subsequent fine-tuning process. It also shows the proportion of time spent applying PEMA relative to fine-tuning. We observe that for five of the nine datasets in GLUE, applying PEMA takes less than half a second. For eight out of nine datasets, applying PEMA takes less than 1\% of the fine-tuning duration. We note that the time taken to apply PEMA correlates with the size of the fine-tuning dataset (Figure~\ref{fig:glue_datatasets}). Overall, we note that the time taken to apply PEMA is generally fractional compared to the fine-tuning duration, even though we made no effort to optimize our implementation. As guidance for future optimization efforts, we note that the dataset processing operations in PEMA are trivially parallelizable.

\newpage

\begin{table}[t]
\scriptsize
\centering
\begin{tabular}{lrrrr}
\toprule
Dataset & \multicolumn{1}{c}{Train} & \multicolumn{1}{c}{Validation} & \multicolumn{1}{c}{Test} & \multicolumn{1}{c}{Total} \\
\midrule
CoLA & 8,551 & 1,043 & 1,063 & 10,657 \\
MNLI & 392,702 & 19,647 & 19,643 & 431,992 \\
MRPC & 3,668 & 408 & 1,725 & 5,801 \\
QNLI & 104,743 & 5,463 & 5,463 & 115,669 \\
QQP & 363,846 & 40,430 & 390,965 & 795,241 \\
RTE & 2,490 & 277 & 3,000 & 5,767 \\
SST-2 & 67,349 & 872 & 1,821 & 70,042 \\
STS-B & 5,749 & 1,500 & 1,379 & 8,628 \\
WNLI & 635 & 71 & 146 & 852 \\
\bottomrule
\end{tabular}
\caption{The number of examples per split in each of the GLUE datasets.}
\label{tab:glue_splits}
\end{table}

\begin{table}[t]
\centering
\scriptsize
\begin{tabular}{lccc}
\toprule
Hyperparameter & GLUE & XNLI \\
\midrule
Adam $\epsilon$ & \multicolumn{2}{c}{1e-8} \\
Adam $\beta_1$ & \multicolumn{2}{c}{0.9} \\
Adam $\beta_2$ & \multicolumn{2}{c}{0.999} \\
Batch Size & \multicolumn{2}{c}{32} \\
Dropout (Attention) & \multicolumn{2}{c}{0.1} \\
Dropout (Hidden) & \multicolumn{2}{c}{0.1} \\
Learning Rate (Peak) & \multicolumn{2}{c}{2e-5, 7.5e-6 (XLM)} \\
Learning Rate Schedule & \multicolumn{2}{c}{Linear} \\
Sequence Length & \multicolumn{2}{c}{128}  \\
Training Epochs & 3 & 2 \\
\bottomrule
\end{tabular}
\caption{The hyperparameters used for each set of experiments.}
\label{tab:hyperparameters}
\end{table}

\begin{table}[t]
\centering
\scriptsize
\begin{tabular}{lrrr}
\toprule
Dataset &\multicolumn{1}{c}{PEMA} & \multicolumn{1}{c}{Fine-tuning} & \multicolumn{1}{c}{\%} \\
\midrule
CoLA & 0.4\sd{0.0} & 172.7\sd{0.9} & 0.2 \\
MNLI & 8.8\sd{0.2} & 7817.8\sd{16.6} & 0.1 \\
MRPC & 0.3\sd{0.0} & 78.7\sd{0.7} & 0.4 \\
QNLI & 2.4\sd{0.0} & 2092.8\sd{2.0} & 0.1 \\
QQP & 13.3\sd{0.5} & 7235.5\sd{4.9} & 0.2 \\
RTE & 0.4\sd{0.0} & 55.4\sd{0.6} & 0.7 \\
SST-2 & 1.2\sd{0.0} & 1329.2\sd{0.3} & 0.1 \\
STS-B & 0.4\sd{0.0} & 118.7\sd{0.5} & 0.3 \\
WNLI & 0.3\sd{0.0} & 18.3\sd{0.8} & 1.4 \\
\bottomrule
\end{tabular}
\caption{The mean duration (seconds) and standard deviation over five runs of applying PEMA to RoBERTa and fine-tuning on the GLUE datasets.}
\label{tab:runtime_impact}
\end{table}

\begin{table*}[t]
\scriptsize
\centering
\begin{tabular}{lcrrrrrrrrrr}
\toprule
Model & PEMA & \multicolumn{1}{c}{CoLA} & \multicolumn{1}{c}{MNLI} & \multicolumn{1}{c}{MRPC} & \multicolumn{1}{c}{QNLI} & \multicolumn{1}{c}{QQP} & \multicolumn{1}{c}{RTE} & \multicolumn{1}{c}{SST-2} & \multicolumn{1}{c}{STS-B} & \multicolumn{1}{c}{WNLI} & \multicolumn{1}{c}{Mean} \\
\midrule
\multirow[c]{2}{*}{DistilBERT} & \ding{55} & 49.3 & 82.2 & 84.2 & 88.5 & 86.7 & 59.6 & 90.5 & 86.5 & 49.3 & 75.2\sd{1.5} \\
 & \ding{51} & 49.3 & 82.2 & 84.2 & 88.6 & 86.7 & 59.6 & 90.5 & 86.5 & 49.3 & 75.2\sd{1.5} \\
\cmidrule{1-12}
\multirow[c]{2}{*}{DistilRoBERTa} & \ding{55} & 56.4 & 84.2 & 85.0 & 90.9 & 87.2 & 65.7 & 92.3 & 87.2 & 53.0 & 78.0\sd{0.9} \\
 & \ding{51} & 56.4 & 84.2 & 85.0 & 90.9 & 87.2 & 65.7 & 92.3 & 87.2 & 53.0 & 78.0\sd{0.9} \\
\cmidrule{1-12}
\multirow[c]{2}{*}{DistilmBERT} & \ding{55} & 29.7 & 78.3 & 81.8 & 86.7 & 85.8 & 60.9 & 89.1 & 84.4 & 48.2 & 71.6\sd{0.3} \\
 & \ding{51} & 29.6 & 78.3 & 81.8 & 86.7 & 85.8 & 60.9 & 89.2 & 84.4 & 48.2 & 71.6\sd{0.4} \\
\cmidrule{1-12}
\multirow[c]{2}{*}{BERT} & \ding{55} & 56.4 & 84.3 & 84.3 & 91.1 & 87.9 & 64.4 & 92.6 & 88.1 & 37.7 & 76.3\sd{0.7} \\
 & \ding{51} & 56.7 & 84.3 & 84.3 & 91.3 & 87.8 & 64.4 & 92.5 & 88.1 & 37.7 & 76.3\sd{0.8} \\
\cmidrule{1-12}
\multirow[c]{2}{*}{RoBERTa} & \ding{55} & 57.6 & 87.8 & 88.4 & 92.8 & 88.4 & 71.1 & 94.2 & 89.9 & 52.1 & 80.3\sd{1.2} \\
 & \ding{51} & 57.6 & 87.8 & 88.4 & 92.7 & 88.4 & 71.1 & 94.2 & 89.9 & 52.1 & 80.3\sd{1.2} \\
\cmidrule{1-12}
\multirow[c]{2}{*}{DeBERTaV3} & \ding{55} & 67.4 & 90.2 & 88.5 & 93.9 & 89.9 & 79.8 & 95.6 & 90.9 & 53.0 & 83.2\sd{0.8} \\
 & \ding{51} & 67.4 & 90.2 & 88.3 & 93.9 & 89.9 & 79.8 & 95.5 & 90.9 & 53.0 & 83.2\sd{0.8} \\
\cmidrule{1-12}
\multirow[c]{2}{*}{mBERT} & \ding{55} & 35.3 & 82.3 & 85.8 & 91.1 & 87.1 & 69.0 & 91.0 & 88.0 & 53.0 & 75.8\sd{2.0} \\
 & \ding{51} & 35.4 & 82.2 & 85.8 & 91.1 & 87.2 & 69.0 & 90.8 & 88.0 & 53.0 & 75.8\sd{2.0} \\
\cmidrule{1-12}
\multirow[c]{2}{*}{XLM-RoBERTa} & \ding{55} & 22.6 & 83.9 & 76.9 & 89.5 & 86.9 & 57.3 & 92.2 & 84.2 & 52.1 & 71.7\sd{2.0} \\
 & \ding{51} & 22.4 & 84.0 & 76.8 & 89.5 & 86.8 & 57.3 & 92.0 & 84.2 & 52.1 & 71.7\sd{2.0} \\
\cmidrule{1-12}
\multirow[c]{2}{*}{XLM-V} & \ding{55} & 0.0 & 84.5 & 68.8 & 89.6 & 86.7 & 54.1 & 91.8 & 80.8 & 55.2 & 68.0\sd{0.6} \\
 & \ding{51} & 0.0 & 84.5 & 68.8 & 89.6 & 86.7 & 54.1 & 91.6 & 80.8 & 55.2 & 67.9\sd{0.6} \\
\bottomrule
\end{tabular}
\caption{Results on the validation set for each task from GLUE. We present the mean performance over five different seeds, accompanied by the overall mean and standard deviation. We report Matthews correlation for CoLA, F1 for QQP, Spearman correlation for STS-B, and accuracy for the remaining tasks. }
\label{tab:glue_performance}
\end{table*}

\end{document}